\documentclass[runningheads]{llncs}

\usepackage{eccv}

\usepackage{eccvabbrv}

\usepackage{graphicx}
\usepackage{booktabs}
\usepackage{algorithm}
\usepackage{algpseudocode}

\usepackage[accsupp]{axessibility}  

\usepackage{hyperref}

\usepackage{orcidlink}

\begin{document}

\title{Learn Once, Edit Anywhere: Visual Direction Transfer for Diffusion Models}

\titlerunning{ViDiT}

\author{Yusuf Dalva\orcidlink{0000-0002-8402-8291} \and
Hidir Yesiltepe\orcidlink{0009-0005-4996-2344} \and
Pinar Yanardag\orcidlink{0009-0003-3452-7417}}

\authorrunning{Y.~Dalva et al.}

\institute{Virginia Tech \\
\email{\{ydalva, hidir, pinary\}@vt.edu}\\
\url{http://vidit-edit.github.io}}

\maketitle
\begin{figure}
    \centering
    \includegraphics[width=\linewidth]{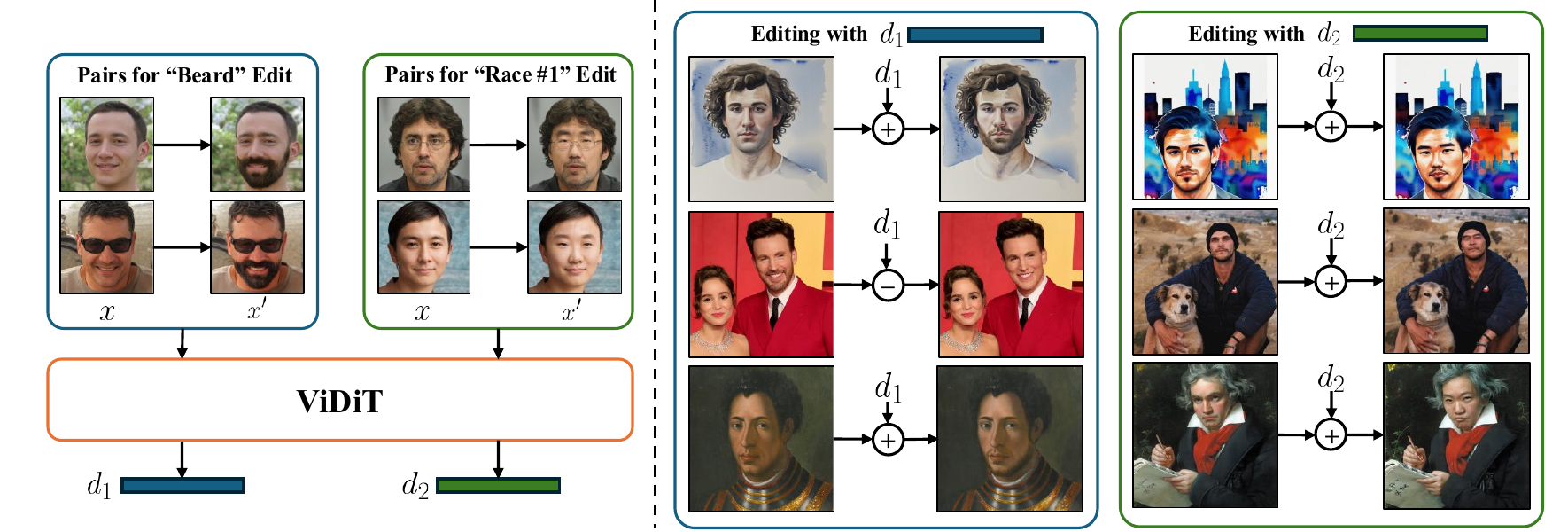}
    \caption{\textbf{ViDiT.} By learning a continuous direction from a small set of image-edit
pairs, ViDiT transfers fine-grained semantic edits to any image in a zero-shot manner.
Once learned, each direction enables disentangled, bidirectional attribute control,
applied with scale $\lambda_e$, across diverse domains including real-world
portraits, artistic styles, and illustrations, without any fine-tuning or per-image
optimization.}
    \label{fig:teaser}
\end{figure}

\begin{abstract}
    The rapid advancement of diffusion models has enabled the generation of high-fidelity images from textual prompts, yet achieving precise, disentangled control over specific attributes remains a significant challenge. A fundamental limitation arises because visual differences between images are often far more descriptive and nuanced than what can be captured through human-crafted text descriptions, which frequently fail to convey fine-grained semantic details. To address this, we introduce \textbf{ViDiT (Visual Direction Transfer for Diffusion)}, a framework that expands the editing vocabulary by capturing latent semantics directly from image-edit pairs. ViDiT learns the underlying transformation by optimizing a single, global, and continuous editing direction from a small set of ``before-and-after'' examples. This optimization process transfers visual changes into the diffusion model's conditioning space, allowing for detailed edits that text alone cannot easily describe. ViDiT operates on a ``Learn Once'' principle, which completely eliminates the need for model fine-tuning or expensive per-image optimization during inference. Once learned, these continuous directions enable ``Edit Anywhere'' capabilities, allowing users to apply highly disentangled manipulations, such as changes in facial features, animal attributes, or artistic styles, to any image in a zero-shot manner with granular control over the edit intensity. Quantitative and qualitative evaluations demonstrate that ViDiT outperforms existing text-based editing methods in maintaining input faithfulness while achieving precise, scalable attribute control.
\end{abstract}
    
\section{Introduction}
\label{sec:intro}

Denoising Diffusion Models (DDMs) \cite{ho2020denoising} and Latent Diffusion Models
(LDMs) \cite{rombach2022high} have gained significant popularity in the generative
modeling landscape due to their ability to synthesize high-quality, high-resolution
images across diverse domains \cite{rombach2022high, ho2020denoising}. Their success
has led researchers to increasingly leverage them for image editing tasks, ranging from
text-driven modifications \cite{hertz2022prompt} to layout-based control
\cite{zhang2023adding}. However, a fundamental limitation persists in these workflows:
human-crafted text descriptions are often too coarse to capture the nuanced visual
differences, or ``deltas'', that define precise, fine-grained edits. While a pair of
images can effortlessly demonstrate a subtle semantic shift, translating that same
transformation into a textual prompt is inherently constrained by the descriptive
bottleneck of natural language. This creates a significant gap between the user's intent
and the model's output, as the expressive power of language is frequently surpassed by
the complexity of visual information.

This gap is further widened by the fact that diffusion models often remain black boxes,
limiting the fine-grained interpretability required for disentangled control. Advancing
such control requires understanding their semantic structures, particularly how specific
attributes can be manipulated in isolation. In this regard, Generative Adversarial
Networks (GANs) have proven to be useful models for achieving interpretable and disentangled edits due
to their well-structured latent spaces that offer rich, linearly applicable semantics.
By leveraging this organized architecture, specialized discovery frameworks
\cite{yuksel2021latentclr, harkonen2020ganspace} have uncovered significant amounts of
semantically meaningful directions \cite{wu2020stylespace}, allowing for highly nuanced
control. In contrast, current research on diffusion models has only uncovered a handful
of disentangled latent directions \cite{dalva2023noiseclr, kwon2022diffusion}. This
disparity arises from the recursive noise estimation and the inherently complex
management of variables across multiple timesteps in diffusion architectures, making it
difficult to map the fine-grained visual deltas that are easily represented in GANs.

To overcome these limitations, we introduce \textbf{ViDiT} (Visual Direction Transfer
for Diffusion), a framework that expands the existing editing vocabulary by capturing
latent semantics directly from image-edit pairs. Operating on a ``Learn Once, Edit
Anywhere'' principle, ViDiT moves away from a total dependency on human-crafted text
by learning to capture latent semantics from a small set of before-and-after visual
examples. Rather than a simple feedforward mapping, ViDiT learns the underlying
transformation by optimizing a single, global, and \textbf{continuous editing
direction} that transfers the visual delta into the diffusion model's conditioning
space. This allows for detailed attribute control that text alone cannot easily
describe, all without requiring base-model finetuning or per-sample test-time
optimization.

Our main contributions are outlined as follows:
\begin{itemize}
    \item We present a universal framework that learns \textbf{continuous editing
    directions} from visual examples, effectively expanding the editing vocabulary of
    pre-trained diffusion models without requiring base model finetuning.

    \item We demonstrate the versatility of our direction transfer by successfully
    drawing from diverse sources, including unconditional generative spaces (e.g.,
    StyleSpace \cite{wu2020stylespace}), domain adaptation models (e.g.,
    StyleGAN-NADA \cite{gal2021stylegannada}), and diffusion-based editors (e.g.,
    Prompt2Prompt \cite{hertz2022prompt}).

    \item We showcase the \textbf{``Edit Anywhere''} capability of our approach by
    transferring a wide range of fine-grained semantics across diverse categories,
    including human faces, artistic styles, and extending these to real-world,
    in-the-wild images.

    \item Our evaluations demonstrate that ViDiT achieves superior disentanglement
    and identity preservation compared to state-of-the-art text-based diffusion
    editors, while providing granular control over edit intensity.

    \item We share our source code to facilitate further research in visual direction transfer for diffusion models.
\end{itemize}

\section{Related Work}

\begin{figure}[t!]
    \centering
    \includegraphics[width=\linewidth]{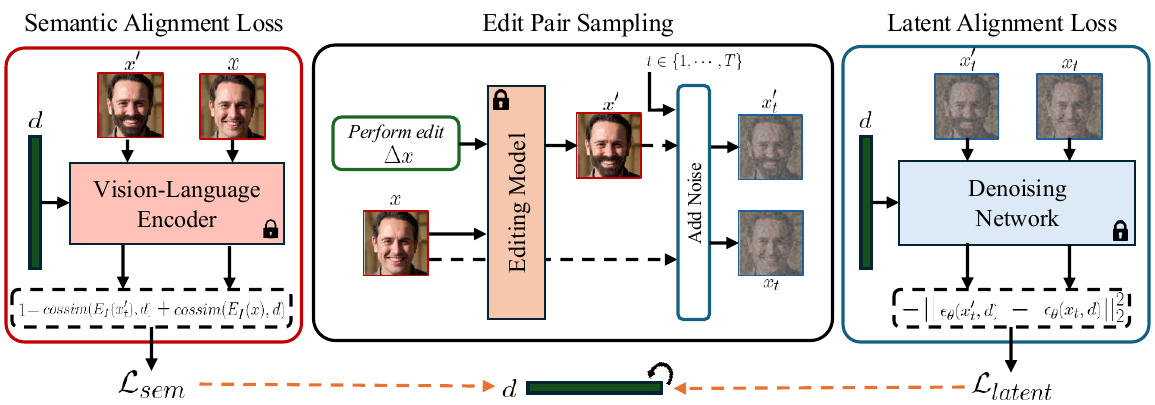}
    \caption{\textbf{ViDiT Framework.}
    Given a pair of images $(x, x')$, where $x' = x + \Delta x$ is obtained by
    applying an edit $\Delta x$ using any editing method, we optimize a continuous
    latent direction $\mathbf{d}$ using two complementary objectives. The
    \textbf{Semantic Alignment Loss} $\mathcal{L}_\text{sem}$ encourages $\mathbf{d}$
    to be geometrically aligned with $x'$ and misaligned with $x$ in the feature space
    of a frozen Vision-Language Encoder ($E$). The \textbf{Latent Alignment Loss}
    $\mathcal{L}_\text{latent}$ maximizes the difference in the denoising predictions
    of the frozen Denoising Network ($\epsilon_\theta$) between the noised edited image
    $x'_t$ and the noised original $x_t$, grounding $\mathbf{d}$ in the diffusion
    model's native latent space. Both losses jointly update $\mathbf{d}$, while all
    other components remain frozen. Once learned, $\mathbf{d}$ can be applied
    zero-shot to any image at inference time.}
    \label{fig:framework}
\end{figure}

\paragraph{Latent Space Exploration of Diffusion Models.}
Text-to-image diffusion models utilize pre-trained text encoders to represent
semantically rich information within their latent spaces, enabling high-quality image
generation from textual conditions~\cite{rombach2022high}. Recent research has sought
to uncover the underlying semantic structure of this feature space to enable more
precise control. Some approaches focus on specific architectural components, such as
the representations encoded in the bottleneck block of the denoising
network~\cite{kwon2022diffusion}, while others investigate the transformation of
representations across different imaging domains~\cite{wu2023latent}. Inspired by
latent space exploration in GANs, researchers have attempted to identify directions
targeted by specific input semantics~\cite{park2023understanding}; however, these
methods often struggle to generalize to large-scale models like Stable Diffusion, as
the diffusion latent space lacks the compact, linear structure that makes GAN spaces
amenable to direction discovery. Recent efforts have addressed this by decomposing
latent variables into energy functions of semantic significance~\cite{Liu_2023_ICCV}
or using contrastive learning to discover unsupervised disentangled
directions~\cite{dalva2023noiseclr}. Despite these advances, such methods remain
constrained by the inherent complexity of the diffusion latent space and are limited
to directions already present within the model's learned representation. \textbf{ViDiT}
bridges this gap by transferring well-understood semantics from diverse external sources
directly into accessible, continuous editing directions.

\paragraph{Editing Visual Semantics with Diffusion Models.}
Editing with diffusion models typically relies on text prompts or structural guidance to
modify attributes~\cite{hertz2022prompt, zhang2023adding}. While effective, these
methods are often limited by the descriptive bottleneck of natural language. SEGA
\cite{brack2023sega} utilizes semantic guidance to steer the diffusion process without
additional training, while InstructPix2Pix~\cite{brooks2023instructpix2pix} learns to
follow human-edited instructions through a conditional diffusion model. Structure-based
methods such as PnP-Diffusion~\cite{tumanyan2023plug} and
MasaCtrl~\cite{cao2023masactrl} inject intermediate features to preserve layout and
identity during editing, and Asyrp~\cite{kwon2022diffusion} and
DiffAE~\cite{preechakul2022diffusion} explore semantic latent spaces within the
denoising architecture itself. Recent works have also explored diverse conditioning
strategies: IP-Adapter~\cite{ye2023ip} injects image-based prompts via decoupled
cross-attention, while Weights2Weights~\cite{dravidinterpreting} and
Attribute-Control~\cite{baumann2025attributecontrol} identify linear directions in the
weight or CLIP embedding spaces, respectively. However, all of these approaches either
depend on coarse textual descriptions, require large-scale training data, or are
restricted to directions expressible through text-level features alone. \textbf{ViDiT}
bridges these paradigms by learning the desired edit once from visual examples and
applying it as a global, \textbf{continuous editing direction}, an approach that
captures attribute-level semantics that are self-evident in visual pairs but remain
difficult for textual or structural instructions to define precisely.

\paragraph{Combining Structured Generative Models and Diffusion.}
Recent literature has explored hybridizing the structured control of GANs with the
generative power of diffusion models~\cite{song2024stylegan}. More relevant to our
work, $\mathcal{W}+$ Adapter~\cite{li2023w-plus-adapter} and Concept
Sliders~\cite{gandikota2023sliders} leverage the disentangled capabilities of
structured latent spaces to enhance diffusion-based editing. Concept
Sliders~\cite{gandikota2023sliders} utilize low-rank adapters (LoRAs) to learn
specific concepts, often employing samples from a pre-trained StyleGAN as a source for
semantic directions, but require training a separate LoRA per concept, which is again bounded by a text prompt. The $\mathcal{W}+$
Adapter~\cite{li2023w-plus-adapter} introduces an additional architectural module to map
StyleGAN's $W+$ latent space directly to the diffusion model, enabling the transfer of
GAN-based edits to individual images. Similarly, DiffAE~\cite{preechakul2022diffusion}
learns a semantic encoder alongside the diffusion decoder to enable structured latent
manipulation, though this requires training a dedicated architecture from scratch. In
contrast, \textbf{ViDiT} eliminates the need for both model fine-tuning and the
introduction of additional adapter modules. By learning a single, global, and
\textbf{continuous editing direction} directly from visual deltas, our framework enables
a ``Learn Once, Edit Anywhere'' capability applicable to any image in a zero-shot
manner. This bypasses the overhead of training per-concept LoRAs or specialized latent
mapping layers, providing a more lightweight and scalable solution for incorporating
complex visual semantics into pre-trained diffusion models.

\section{Method}
\label{sec:method}

\begin{figure}[t!]
    \centering
    \includegraphics[width=\linewidth]{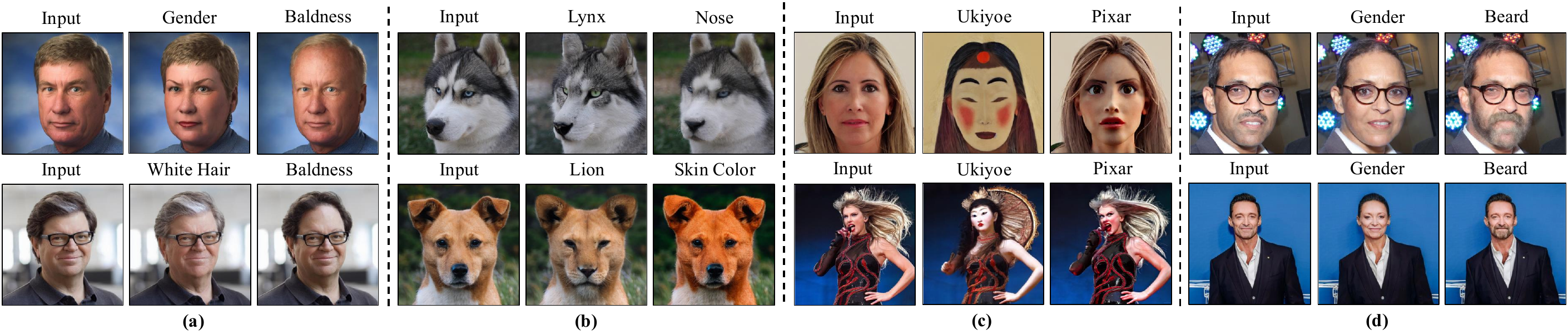}
    \caption{\textbf{Directions transferred with ViDiT.}
    ViDiT transfers fine-grained editing directions from diverse supervision sources
    to in-the-wild images in a zero-shot manner.
    \textbf{(a)} Directions transferred from StyleGAN2~\cite{karras2020analyzing}
    trained on FFHQ~\cite{karras2019style}, demonstrating facial attribute edits such
    as \textit{Bald}, \textit{Gender}, \textit{White Hair}, and \textit{Baldness}.
    \textbf{(b)} Directions transferred from StyleGAN2 trained on
    AFHQ~\cite{choi2020starganv2}, showcasing animal attribute edits including
    \textit{Lynx}, \textit{Nose}, \textit{Lion}, and \textit{Skin Color}.
    \textbf{(c)} Style directions transferred from
    StyleGAN-NADA~\cite{gal2021stylegannada}, applying artistic transformations such
    as \textit{Ukiyoe} and \textit{Pixar} to diverse real-world images.
    \textbf{(d)} Semantic directions transferred from
    Prompt2Prompt~\cite{hertz2022prompt}, enabling attribute edits such as
    \textit{Gender} and \textit{Beard} on real portraits.
    Across all cases, edits are disentangled and faithfully preserve the identity,
    background, and artistic style of the input.}
    \label{fig:main-results}
\end{figure}

\subsection{Preliminaries}

\noindent\textbf{Diffusion Models.}
Diffusion models~\cite{ho2020denoising, rombach2022high} learn to generate data through
an iterative reverse denoising process. Given a clean image $x_0$, the forward process
adds Gaussian noise $\epsilon \sim \mathcal{N}(0, 1)$ over $T$ timesteps to produce a
noisy latent $x_t$. A denoising network $\epsilon_\theta$ is trained to predict this
noise via the objective:
\begin{equation}
    \label{eqn:denoising_network}
    \mathcal{L}_{DM} = \mathbb{E}_{x_0, \epsilon, t} [ ||\epsilon -
    \epsilon_\theta(x_t, t)||^2_2 ]
\end{equation}
During inference, sampling begins at $x_T \sim \mathcal{N}(0, 1)$ and iteratively
approaches $x_0$. To steer this process, Classifier-Free Guidance
(CFG)~\cite{ho2022classifier} computes a weighted combination of conditional and
unconditional noise predictions:
\begin{equation}
    \label{eqn:cfg}
    \tilde{\epsilon}_\theta(x_t, c) = \epsilon_\theta(x_t, \phi) + \lambda_g
    (\epsilon_\theta(x_t, c) - \epsilon_\theta(x_t, \phi))
\end{equation}
where $c$ is a text-derived conditioning embedding, $\phi$ is the null-text embedding,
and $\lambda_g$ is the guidance scale. A key observation is that the CFG residual
$(\epsilon_\theta(x_t, c) - \epsilon_\theta(x_t, \phi))$ acts as a \emph{directional
signal} in the noise-prediction space that steers generation towards $c$. ViDiT
exploits this structure by learning a continuous direction $\mathbf{d}$ that replaces
the text conditioning signal entirely, allowing semantic edits to be expressed as
directions in this space rather than as natural language descriptions.

\subsection{ViDiT: Transferring Visual Deltas}

\noindent\textbf{Motivation.}
Existing text-driven editing methods rely on natural language to specify semantic
changes~\cite{brack2023sega, brooks2023instructpix2pix, hertz2022prompt}. While
convenient, this imposes a \emph{linguistic bottleneck}: many fine-grained visual
concepts, precise facial structure, subtle texture shifts, nuanced style changes,
are self-evident in images but extremely difficult to describe in text. A prompt like
``a person with a beard'' captures only a coarse prior and inevitably entangles the edit
with unrelated attributes such as age or skin tone. By contrast, a set of
before-and-after image pairs $\{(x, x')\}$, where $x' = x + \Delta x$ is obtained by
applying an edit $\Delta x$ through any editing method, encodes the precise visual delta
without linguistic ambiguity. Our key insight is that this delta can be
\emph{transferred} into a compact, continuous direction $\mathbf{d}$ that lives in the
diffusion model's conditioning space and faithfully replicates the transformation on any
new image.

\noindent\textbf{Overview.}
ViDiT identifies a continuous latent direction $\mathbf{d}$, which acts as a
transferable visual prompt, by optimising it against a small set of $N$ image-edit pairs
$\{\mathcal{X}_\text{input}, \mathcal{X}_\text{edited}\}$. The direction $\mathbf{d}$
is initialised randomly within the model's text-embedding space and refined using a
dual-objective loss that enforces alignment at two complementary levels of abstraction:
globally, via semantic feature matching, and locally, via dense latent-space
correspondence. Crucially, the denoising network $\epsilon_\theta$ and the
vision-language encoder $E_I$ are both kept frozen throughout; only $\mathbf{d}$ is
updated. This design ensures that the model's generative prior is not disturbed and that
$\mathbf{d}$ genuinely captures the target transformation rather than overfitting to
the specific images seen during training.

\paragraph{Semantic Alignment Loss.}
To provide a global semantic anchor, we leverage a frozen Vision-Language
Encoder~$E_I$~\cite{radford2021learning}. For $\mathbf{d}$ to represent the
intended semantic shift, it must be geometrically close to the edited images in the
feature space of the encoder, while remaining distant from the originals. We formulate
$\mathcal{L}_\text{sem}$ as a contrastive objective:
\begin{equation}
\label{eqn:sem-align}
    \mathcal{L}_\text{sem} = 1 - \text{cossim}(E_I(x'), \mathbf{d}) +
    \text{cossim}(E_I(x), \mathbf{d})
\end{equation}
This loss places $\mathbf{d}$ in the correct conceptual neighbourhood of the target
concept. However, because CLIP operates on globally pooled, high-level embeddings, it
is insensitive to fine-grained spatial structure. A direction satisfying
$\mathcal{L}_\text{sem}$ alone may capture the right concept yet still alter unrelated
facial features, since CLIP cannot distinguish between a change in beard texture and a
correlated change in jaw shape. This motivates our second loss term.

\paragraph{Latent Alignment Loss.}
To capture high-frequency structural details and enforce pixel-level correspondence
between pairs, we introduce $\mathcal{L}_\text{latent}$. Unlike GANs, which condition
on compact low-dimensional latent codes, diffusion models produce spatially-aligned,
high-dimensional noise-prediction maps, $\epsilon_\theta(x_t, \mathbf{d})$, that encode dense per-pixel information. We exploit
this property by maximising the difference between noise predictions for the noised
edited image $x_t'$ and the noised original $x_t$ at random timesteps
$t \sim \mathcal{U}(1, T)$:
\begin{equation}
    \label{eqn:latent-align}
    \mathcal{L}_\text{latent} = -\mathbb{E}_{x_0, \epsilon, t}
    [ ||\epsilon_\theta(x_t', \mathbf{d}) - \epsilon_\theta(x_t, \mathbf{d})||_2^2 ]
\end{equation}
Intuitively, this encourages $\mathbf{d}$ to produce predictions that maximally
distinguish the edited from the unedited image in the denoising network's internal
feature space, grounding the direction in the model's native representation. In
practice, we observe that $\mathcal{L}_\text{latent}$ is critical for preserving
fine-grained identity cues such as hair follicle structure, specular reflections on
glasses, and skin texture, details that $\mathcal{L}_\text{sem}$ alone cannot
recover. The two losses are complementary: $\mathcal{L}_\text{sem}$ steers $\mathbf{d}$
towards the correct semantic concept, while $\mathcal{L}_\text{latent}$ sharpens its
spatial precision. The total training objective is $\mathcal{L} = \mathcal{L}_\text{sem} + \mathcal{L}_\text{latent}$. We provide the training algorithm for ViDiT in Alg. \ref{alg:training}.

\begin{algorithm}[t!]
    \caption{Learning direction $\mathbf{d}$ with ViDiT}\label{alg:training}
    \begin{algorithmic}
        \Require Pre-trained diffusion model $\epsilon_\theta(x_t, c)$, pre-trained
        CLIP Image Encoder $E_I(x)$, a set of $N$ image-edit pairs
        $\{(x_1, x_1'), \cdots, (x_N, x_N')\}$ obtained from any editing source,
        randomly initialized direction $\mathbf{d}$, learning rate $\eta$.
        \While{training}
        \For{$i = 1, \cdots, N$}
        \State Sample pair $(x_i, x_i')$
        \State Sample $\epsilon \sim \mathcal{N}(0,1)$, $t \sim \mathcal{U}(1, T)$
        \State $x_{i,t} = x_i + \alpha^t \epsilon$,\quad $x_{i,t}' = x_i' + \alpha^t \epsilon$
        \State $\mathcal{L}_\text{latent} = -||\epsilon_\theta(x_{i,t}', \mathbf{d}) - \epsilon_\theta(x_{i,t}, \mathbf{d})||_2^2$
        \State $\mathcal{L}_\text{sem} = 1 - \mathrm{cossim}(E_I(x_i'), \mathbf{d}) + \mathrm{cossim}(E_I(x_i), \mathbf{d})$
        \State $\mathcal{L} = \mathcal{L}_\text{sem} + \mathcal{L}_\text{latent}$
        \State $\mathbf{d} \leftarrow \mathbf{d} - \eta \nabla_{\mathbf{d}} \mathcal{L}$
        \EndFor
        \EndWhile
    \end{algorithmic}
\end{algorithm}

\subsection{The ``Edit Anywhere'' Principle}

Once $\mathbf{d}$ has been optimized, a one-time process per editing concept that
requires no fine-tuning of the base model, it effectively expands the diffusion
model's vocabulary with a new, reusable semantic token. At inference time, $\mathbf{d}$
can be injected into the CFG formulation to steer the denoising trajectory of
\emph{any} image towards the target concept, regardless of domain, style, or content.

\noindent\textbf{Editing via Direction Injection.}
We modify the inference-time CFG to incorporate the learned direction:
\begin{equation}
    \label{eqn:editing}
    \bar{\epsilon}_\theta(x_t, c, \mathbf{d}) = \tilde{\epsilon}_\theta(x_t, c)
    + \lambda_e (\epsilon_\theta(x_t, \mathbf{d}) - \epsilon_\theta(x_t, \phi))
\end{equation}
where the additional term replaces the standard text-conditioning residual with the
visual-delta residual, and $\lambda_e \in \mathbb{R}$ provides continuous, granular
control over edit intensity. Setting $\lambda_e = 0$ recovers the unedited output,
while increasing $\lambda_e$ progressively strengthens the effect. Crucially, this
requires no per-image optimization or test-time adaptation: the same $\mathbf{d}$
applies zero-shot to any input.

\paragraph{Multi-directional Editing.}
A key advantage of the additive formulation in Eq.~\ref{eqn:editing} is that multiple
independently learned directions can be composed at inference time without retraining.
Given a set of directions $D = \{d_1, \dots, d_k\}$ with corresponding editing scales
$\{\lambda_{e_1}, \dots, \lambda_{e_k}\}$, the noise prediction generalizes to:
\begin{equation}
    \label{eqn:multi-edit}
    \hat{\epsilon}_\theta(x_t, c, D) = \tilde{\epsilon}_\theta(x_t, c) +
    \sum_{i=1}^{|D|} \lambda_{e_i} (\epsilon_\theta(x_t, d_i) -
    \epsilon_\theta(x_t, \phi))
\end{equation}
This allows complex combinations of attributes, for example, simultaneously applying
a smile direction and an age direction, while preserving strong disentanglement
between each component, since each $d_i$ was optimized independently to isolate a
single semantic concept.

\paragraph{Real Image Editing.}
ViDiT directions generalize robustly to real-world photographs. To edit a real image,
we first invert it into the model's latent trajectory using DDPM
Inversion~\cite{huberman2023edit}, which produces a sequence of noisy latents
$\{x_T, \dots, x_0\}$ that approximately reconstruct the input when denoised. We then
perform the reverse process with the direction-injected noise prediction:
\begin{equation}
    \label{eqn:real-edit}
    \bar{\epsilon}_\theta(x_t, \mathbf{d}) = \epsilon_\theta(x_t, \phi)
    + \lambda_e (\epsilon_\theta(x_t, \mathbf{d}) - \epsilon_\theta(x_t, \phi))
\end{equation}
The success of this zero-shot transfer to real images confirms that $\mathbf{d}$ encodes
a fundamental semantic vector within the diffusion model's latent space, rather than an
overfitted representation tied to the distribution of training pairs. As demonstrated in
\cref{fig:main-results}, the learned directions transfer faithfully across synthetic
portraits, animal photographs, and stylized artwork alike.

\section{Experiments}
\label{sec:experiments}

To evaluate the effectiveness of ViDiT in transferring semantically rich latent
directions, we conduct experiments across human faces, cats, and dogs. We demonstrate
that by transferring visual deltas into a continuous direction $\mathbf{d}$, our
framework achieves a ``Learn Once, Edit Anywhere'' capability that generalises across
synthetic, real-world, and stylised images.

\paragraph{Experimental Setup.}
We utilise Stable Diffusion~\cite{rombach2022high} for all experiments. For supervision,
we primarily use off-the-shelf StyleGAN2~\cite{karras2020analyzing} models trained on
FFHQ~\cite{karras2019style}, AFHQ-Cats, and AFHQ-Dogs~\cite{choi2020starganv2}. GANs
are chosen as the primary supervision source because they currently provide the
highest-quality disentangled image-edit pairs; however, ViDiT is source-agnostic and
we additionally demonstrate directions transferred from
StyleGAN-NADA~\cite{gal2021stylegannada} and Prompt2Prompt ~\cite{hertz2022prompt} in
\cref{fig:main-results}. We optimize $\mathbf{d}$ using $N{=}1000$ samples with the
AdamW optimiser~\cite{loshchilov2017decoupled} for 1000 iterations. Once learned,
$\mathbf{d}$ acts as a new ``token'' in the model's vocabulary, enabling zero-shot
editing in ${\sim}5$ seconds on a single NVIDIA L40 GPU.

\subsection{Ablation Studies}
\label{sec:ablations}

\begin{figure}[t!]
    \centering
    \includegraphics[width=0.65\linewidth]{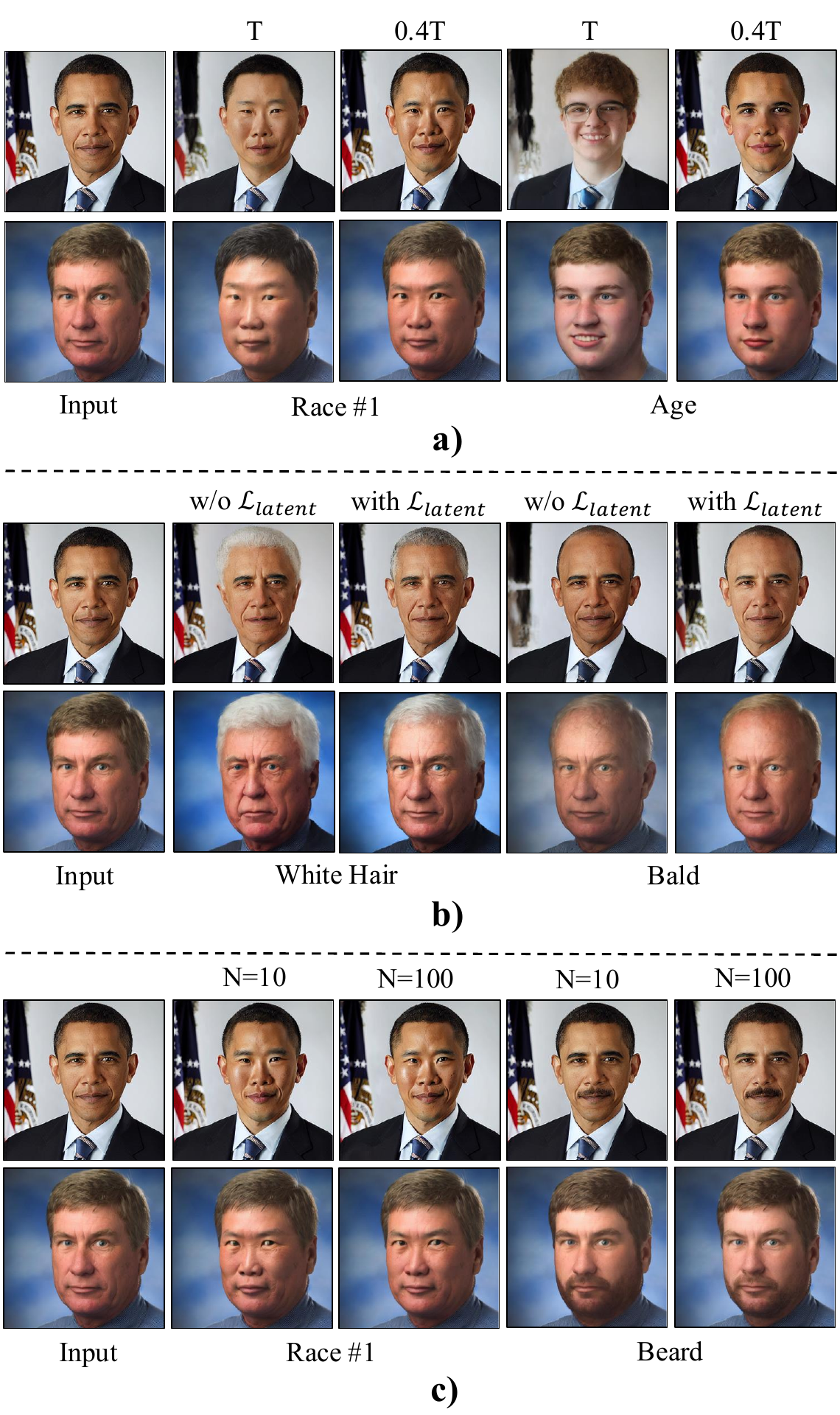}
    \caption{%
        \textbf{Qualitative Ablation Results.}
        \textbf{(a)} Effect of inference timestep $T$: applying the edit at a reduced
        timestep ($0.4T$) yields stronger edits at the cost of greater identity
        deviation, while the full timestep $T$ produces more conservative,
        identity-preserving results.
        \textbf{(b)} Impact of $\mathcal{L}_\text{latent}$: without it, edits exhibit
        entanglement and unintended structural changes (e.g.\ skin tone shift in White
        Hair, over-smoothing in Bald); adding $\mathcal{L}_\text{latent}$ produces
        cleaner, more disentangled edits while preserving identity.
        \textbf{(c)} Effect of training sample size $N$: increasing from $N{=}10$ to
        $N{=}100$ improves edit sharpness and disentanglement, with diminishing returns
        suggesting that meaningful directions can be learned from very few pairs.
    }
    \label{fig:ablations}
\end{figure}

We evaluate four core design choices: the inference timestep $T$, the number of
training pairs $N$, the choice of inversion method, and the contribution of each loss
term. Identity preservation is assessed via LPIPS~\cite{zhang2018unreasonable},
DINO~\cite{oquab2023dinov2}, and DreamSim~\cite{fu2023dreamsim}, and semantic alignment
via CLIP-T~\cite{radford2021learning} and SigLIP-T~\cite{zhai2023sigmoid}.
Qualitative results are shown in \cref{fig:ablations,fig:additional_comps} and
quantitative results in \cref{tab:ablation}.

\noindent\textbf{Ablation on Inference Timestep.}
The injection timestep $T$ controls the trade-off between edit strength and identity
preservation. When $\mathbf{d}$ is applied at the full denoising timestep, the edit is
introduced early in the generative process, allowing the model to gently incorporate the
semantic change while maintaining structural coherence. Reducing to $0.4T$ applies the
direction later, where the denoising network operates on lower-noise latents and produces
more aggressive edits, useful when a strong effect is desired, but at the cost of
noticeable identity drift, as shown in \cref{fig:ablations}(a). In our default
configuration we use the full timestep, which offers the best balance between edit
fidelity and identity preservation.

\noindent\textbf{Ablation on Sample Size.}
A key practical question is how many image-edit pairs are needed to learn a reliable
direction. As shown in \cref{tab:ablation}, ViDiT already captures meaningful semantic
content with as few as $N{=}10$ pairs, demonstrating that the method is not
data-hungry. However, DINO and LPIPS scores improve consistently as $N$ increases, with
$N{=}1000$ yielding the best identity preservation overall. This suggests that a larger
pool of pairs provides broader coverage of the target concept's variation, leading to a
more robustly disentangled direction that generalises better at inference time. Notably,
CLIP-T peaks at $N{=}100$, indicating a mild trade-off between semantic alignment and
structural faithfulness at very large $N$, which we address through our dual-loss
formulation.

\noindent\textbf{Ablation on Inversion Method.}
For real image editing, we ablate the choice of inversion strategy used to map the input
into the model's latent trajectory. As shown in \cref{fig:additional_comps}(b), DDIM
Inversion~\cite{song2020denoising} introduces noticeable drift in fine facial details,
degrading structural consistency between the original and edited outputs. DDPM
Inversion~\cite{huberman2023edit}, by contrast, produces a more faithful latent
trajectory that preserves high-frequency identity cues such as hair texture, glasses
frames, and skin tone, yielding significantly cleaner edits. We therefore adopt DDPM
Inversion as our default for real image editing throughout all experiments.

\noindent\textbf{Ablation on Loss Terms.}
We isolate the contribution of $\mathcal{L}_\text{latent}$ by training with only the
semantic alignment loss $\mathcal{L}_\text{sem}$. While $\mathcal{L}_\text{sem}$ alone
is sufficient to identify the target concept, the resulting direction moves the output
in the correct semantic direction, the absence of $\mathcal{L}_\text{latent}$ leads
to entangled edits that undesirably alter facial structure and skin tone, as visible in
\cref{fig:ablations}(b). This is because $\mathcal{L}_\text{sem}$ operates on globally
pooled CLIP features, which lack the spatial resolution to enforce pixel-level identity
consistency. Adding $\mathcal{L}_\text{latent}$ grounds $\mathbf{d}$ in the denoising
network's dense, spatially-aligned prediction space, yielding a marked improvement in
LPIPS and DINO without sacrificing semantic
alignment, confirming that the two losses are complementary rather than competing.

\begin{table}[t]
  \centering
  \caption{%
    \textbf{Quantitative Ablation Results.}
    Analysis of sample size $N$ and the impact of the Latent Alignment Loss
    ($\mathcal{L}_\text{latent}$). Each variant is evaluated with
    LPIPS~\cite{zhang2018unreasonable}, DINO~\cite{oquab2023dinov2},
    DreamSim~\cite{fu2023dreamsim}, CLIP-T~\cite{radford2021learning},
    and SigLIP-T~\cite{zhai2023sigmoid}.
  }
  \label{tab:ablation}
  \begin{tabular}{lccccc}
    \toprule
    Method & LPIPS$\downarrow$ & CLIP-T$\uparrow$ & DINO$\uparrow$ & SigLIP-T$\uparrow$ & DreamSim$\uparrow$ \\
    \midrule
    $N{=}10$
      & \underline{0.098} & 0.403 & \underline{0.869} & 0.143 & \textbf{0.872} \\
    $N{=}100$
      & 0.136 & \textbf{0.425} & 0.842 & \textbf{0.148} & 0.771 \\
    \midrule
    w/o $\mathcal{L}_\text{latent}$
      & 0.121 & \underline{0.409} & 0.832 & 0.143 & 0.860 \\
    \midrule
    \textbf{Ours (ViDiT)}
      & \textbf{0.093} & 0.406 & \textbf{0.891} & \underline{0.147} & \underline{0.861} \\
    \bottomrule
  \end{tabular}
\end{table}

\subsection{Qualitative Results}
\label{sec:qualitative}

\cref{fig:main-results} demonstrates ViDiT's editing capabilities across a wide range
of semantics and domains. Our method transfers fine-grained directions spanning facial
attributes such as age, gender, beard, and eyeglasses, as well as animal domains
including cats and dogs. A key property of our edits is their high degree of
disentanglement: only the targeted attribute is modified while all other aspects of the
image, including identity, background, and artistic style, are faithfully
preserved. Furthermore, the learned direction $\mathbf{d}$ generalizes robustly beyond
its supervision source: edits optimized on structured GAN pairs apply equally well to
stylized illustrations, classical paintings, and real-world photographs, confirming that
$\mathbf{d}$ encodes a fundamental semantic vector within the diffusion model rather
than an overfitted representation tied to the source domain.

\cref{fig:additional_comps}(c) further demonstrates the compositional power of ViDiT.
Since each direction is learned independently and injected additively at inference time,
multiple directions can be combined simultaneously without retraining, producing coherent
multi-attribute edits such as simultaneously applying \textit{Asian}, \textit{Beard},
and \textit{Smile} while preserving strong disentanglement between each component.

\begin{figure}[t]
    \centering
    \includegraphics[width=\linewidth]{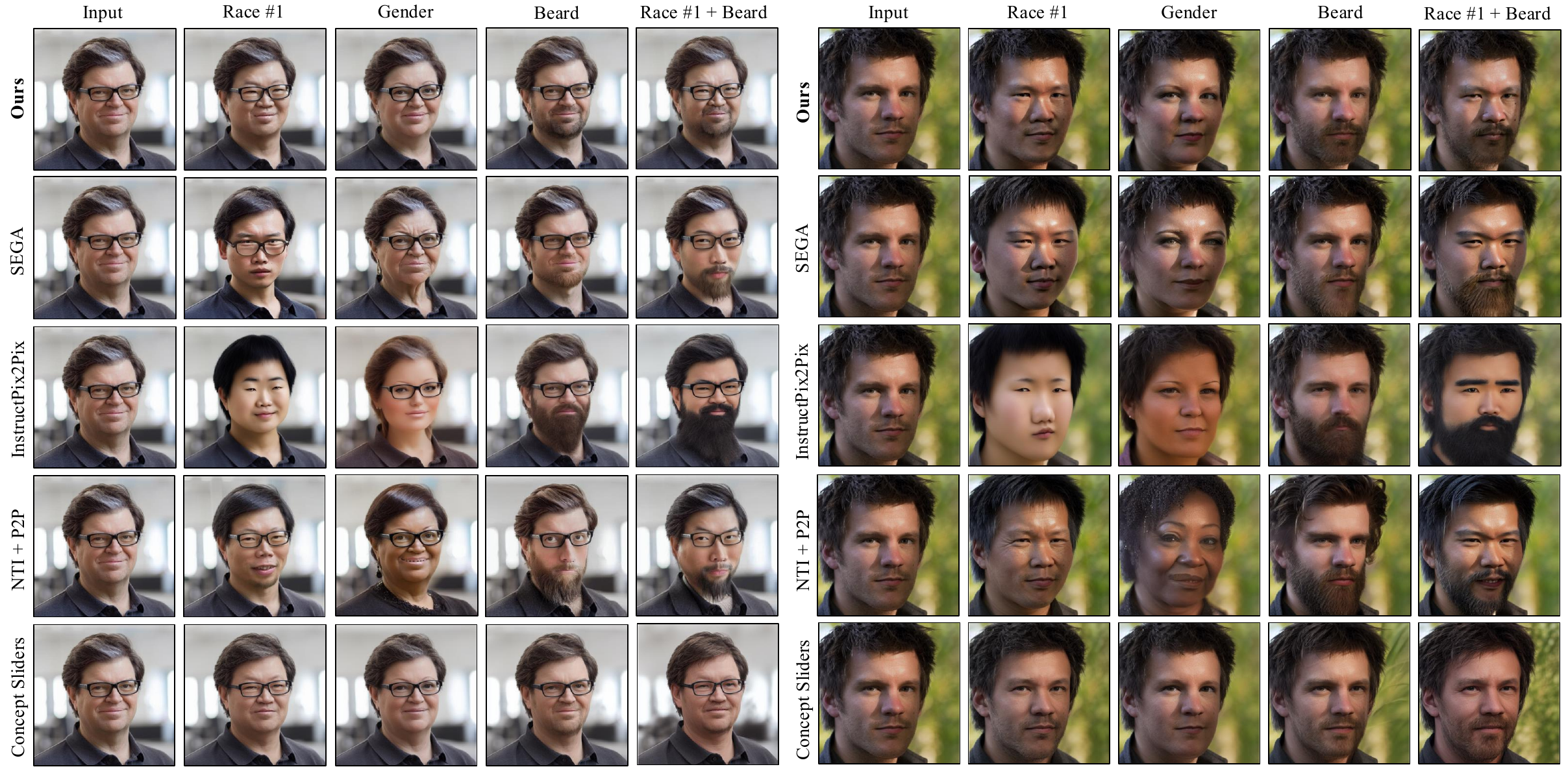}
    \caption{%
        \textbf{Qualitative Comparison with Diffusion-based Image Editing Methods.}
        We compare our approach with Concept Sliders~\cite{gandikota2023sliders},
        SEGA~\cite{brack2023sega}, InstructPix2Pix~\cite{brooks2023instructpix2pix},
        and Prompt2Prompt~\cite{hertz2022prompt}. ViDiT outperforms all baselines in
        achieving disentangled edits across both single semantics and combined
        attributes (e.g.\ ``Race'' and ``Beard''), while faithfully preserving the
        identity and structure of the input.
    }
    \label{fig:quali_comp}
\end{figure}

\subsection{Quantitative Comparisons}
\label{sec:comparison}

\noindent\textbf{Comparison with Visual Conditioning Methods.}
To quantify the advantage of visual-delta supervision over text-based and visual
conditioning alternatives, we evaluate all methods on 100 generated samples. As shown
in \cref{tab:visual_comparison}, text-based and weight-space methods suffer from
significant semantic entanglement, unintentionally altering subject identity to match
coarse linguistic or model priors. By contrast, ViDiT outperforms all competing methods
on identity-preservation metrics (CLIP-I, SigLIP-I, DINO) while remaining competitive
in semantic alignment.

\begin{table}[t]
  \centering
  \caption{%
    \textbf{Comparison with Visual Conditioning and Text-based Methods.}
    Evaluated on 100 generated samples using CLIP-T~\cite{radford2021learning},
    SigLIP-T~\cite{zhai2023sigmoid}, CLIP-I, SigLIP-I, and DINO~\cite{oquab2023dinov2}.
  }
  \label{tab:visual_comparison}
  \begin{tabular}{lccccc}
    \toprule
    Method & CLIP-T$\uparrow$ & SigLIP-T$\uparrow$ & CLIP-I$\uparrow$ & SigLIP-I$\uparrow$ & DINO$\uparrow$ \\
    \midrule
    Text Prompt
      & \textbf{0.405} & 0.135 & 0.775 & 0.875 & 0.822 \\
    Weights2Weights~\cite{dravidinterpreting}
      & 0.401 & \underline{0.146} & 0.686 & 0.837 & 0.642 \\
    IP-Adapter~\cite{ye2023ip}
      & 0.394 & \textbf{0.148} & 0.816 & 0.890 & 0.767 \\
    Attribute-Control~\cite{baumann2025attributecontrol}
      & 0.356 & 0.129 & 0.829 & 0.911 & 0.881 \\
    W+ Adapter~\cite{li2023w-plus-adapter}
      & 0.355 & 0.130 & 0.723 & 0.824 & 0.759 \\
    \midrule
    \textbf{Ours (ViDiT)}
      & \underline{0.395} & 0.143 & \textbf{0.831} & \textbf{0.913} & \textbf{0.886} \\
    \bottomrule
  \end{tabular}
\end{table}

\begin{figure}[t]
    \centering
    \includegraphics[width=\linewidth]{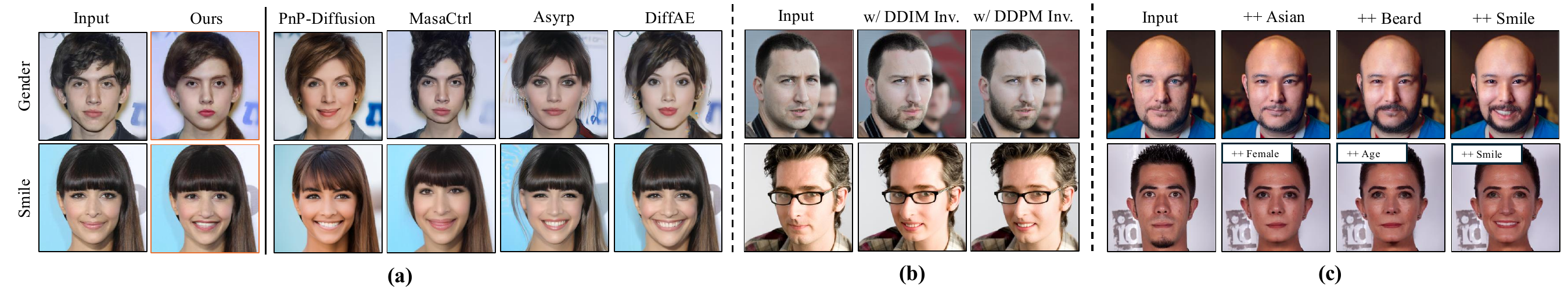}
    \caption{%
        \textbf{Additional Qualitative Results.}
        \textbf{(a)} Extended comparison with diffusion-based editing methods on
        \textit{Gender} and \textit{Smile} semantics. ViDiT produces more
        identity-preserving and disentangled edits compared to
        PnP-Diffusion~\cite{tumanyan2023plug}, MasaCtrl~\cite{cao2023masactrl},
        Asyrp~\cite{kwon2022diffusion}, and DiffAE~\cite{preechakul2022diffusion},
        which tend to alter unrelated facial attributes or global image structure.
        \textbf{(b)} Ablation on the inversion method: DDPM
        Inversion~\cite{huberman2023edit} yields superior structural consistency and
        identity preservation compared to DDIM Inversion~\cite{song2020denoising},
        which introduces noticeable drift in fine facial details.
        \textbf{(c)} Compositional editing via multiple simultaneous directions.
        Applying \textit{++Asian}, \textit{++Beard}, and \textit{++Smile} (top row),
        and \textit{++Female}, \textit{++Age}, and \textit{++Smile} (bottom row)
        simultaneously produces coherent, disentangled multi-attribute edits without
        any retraining.
    }
    \label{fig:additional_comps}
\end{figure}

\noindent\textbf{Comparison with Text-driven Editing Methods.}
We benchmark ViDiT against text-driven editing methods, SEGA~\cite{brack2023sega},
Prompt2Prompt~\cite{hertz2022prompt}, InstructPix2Pix~\cite{brooks2023instructpix2pix},
and Concept Sliders~\cite{gandikota2023sliders}, on 200 input-edit pairs across
``Asian'' (Race \#1) and ``Smile'' semantics. Identity preservation is evaluated with
LPIPS ~\cite{zhang2018unreasonable} and DINO~\cite{oquab2023dinov2}, and semantic
alignment with CLIP-T~\cite{radford2021learning} and
SigLIP-T~\cite{zhai2023sigmoid}. Importantly, our edits are applied without access to
any target text prompt, relying solely on the direction transferred from visual pairs.
As shown in \cref{tab:comparison}, ViDiT significantly outperforms all text-driven
baselines on identity preservation while remaining competitive in semantic alignment.
An extended qualitative comparison against a broader set of baselines, including
PnP-Diffusion~\cite{tumanyan2023plug}, MasaCtrl~\cite{cao2023masactrl},
Asyrp~\cite{kwon2022diffusion}, and DiffAE~\cite{preechakul2022diffusion}, is
provided in \cref{fig:additional_comps}(a), further confirming the superiority of
ViDiT in preserving identity and disentangling targeted attributes.

\begin{table}[t]
  \centering
  \caption{%
    \textbf{Comparison with Text-driven Editing Methods.}
    ViDiT is compared against text-driven baselines using
    LPIPS~\cite{zhang2018unreasonable}, DINO~\cite{oquab2023dinov2},
    CLIP-T~\cite{radford2021learning}, SigLIP-T~\cite{zhai2023sigmoid},
    and DreamSim~\cite{fu2023dreamsim}.
  }
  \label{tab:comparison}
  \begin{tabular}{lccccc}
    \toprule
    Method & LPIPS$\downarrow$ & CLIP-T$\uparrow$ & DINO$\uparrow$ & SigLIP-T$\uparrow$ & DreamSim$\uparrow$ \\
    \midrule
    SEGA~\cite{brack2023sega}
      & 0.179 & 0.388 & 0.714 & 0.134 & 0.757 \\
    Prompt2Prompt~\cite{hertz2022prompt}
      & 0.074 & \textbf{0.408} & 0.867 & \underline{0.143} & 0.869 \\
    InstructPix2Pix~\cite{brooks2023instructpix2pix}
      & 0.059 & 0.403 & 0.851 & \textbf{0.145} & 0.877 \\
    Concept Sliders~\cite{gandikota2023sliders}
      & 0.121 & 0.325 & 0.842 & 0.097 & 0.844 \\
    \midrule
    \textbf{Ours (ViDiT)}
      & \textbf{0.030} & \underline{0.407} & \textbf{0.929} & 0.139 & \textbf{0.905} \\
    \bottomrule
  \end{tabular}
\end{table}

\noindent\textbf{User Study and Rescoring Analysis.}
We conducted a perceptual study with 40 participants on Prolific.com on real images,
comparing ViDiT against SEGA~\cite{brack2023sega},
InstructPix2Pix~\cite{brooks2023instructpix2pix}, Prompt2Prompt~\cite{hertz2022prompt},
and Concept Sliders~\cite{gandikota2023sliders}. Participants rated 60 input-edit pairs
on a 1--5 scale for semantic success and disentanglement quality. As reported in
\cref{tab:userstudy}, ViDiT achieves the highest mean preference score, confirming that
our zero-shot edits are perceived as more disentangled and semantically faithful than
existing text-driven alternatives.

To further assess disentanglement, we perform a rescoring analysis measuring how CLIP
classification probabilities for specific attributes shift following an edit,
following~\cite{shen2020interfacegan, dalva2023noiseclr}. We apply four editing
directions, Asian (Race \#1), Smile, Gender, and Beard, to 100 Stable Diffusion-generated
images and report the resulting shifts in \cref{tab:rescoring}. Targeted edits strongly
increase the probability of the corresponding attribute (diagonal entries), while
off-diagonal interactions remain minimal for unrelated attributes, confirming
disentanglement. Some correlated shifts are semantically expected: the Gender edit
towards femininity notably reduces the Beard score, and the Beard edit slightly
diminishes Smile, consistent with known biases in the underlying generative model.

\begin{table}[t]
  \centering
  \begin{minipage}[t]{0.46\columnwidth}
    \centering
    \caption{%
      \textbf{User Study.}
      Mean preference scores (1--5) from 40 participants on disentanglement
      and semantic faithfulness across 60 input-edit pairs.
    }
    \label{tab:userstudy}
    \small
    \begin{tabular}{lc}
      \toprule
      Method & Preference $\uparrow$ \\
      \midrule
      SEGA~\cite{brack2023sega}                                    & 1.96 \\
      InstructPix2Pix~\cite{brooks2023instructpix2pix}             & 2.06 \\
      Prompt2Prompt~\cite{hertz2022prompt}                         & 2.74 \\
      Concept Sliders~\cite{gandikota2023sliders}                  & 3.25 \\
      \midrule
      \textbf{Ours (ViDiT)}                                        & \textbf{3.36} \\
      \bottomrule
    \end{tabular}
  \end{minipage}
  \hfill
  \begin{minipage}[t]{0.50\columnwidth}
    \centering
    \caption{%
      \textbf{Rescoring Analysis.}
      Shift in CLIP classification probability (\%) per attribute.
      Rows: applied edit; columns: measured attribute.
    }
    \label{tab:rescoring}
    \small
    \begin{tabular}{lcccc}
      \toprule
      & Asian & Smile & Gender & Beard \\
      \midrule
      Asian  & \textbf{53.6} &  15.8  & $-$12.1 & $-$3.5  \\
      Smile  & $-$20.4 & \textbf{41.2}  &  11.8   & $-$7.2  \\
      Gender &   2.0   & $-$4.3 & \textbf{94.7}   & $-$19.8 \\
      Beard  & $-$2.6  & $-$8.1 & $-$0.05 & \textbf{28.3}  \\
      \bottomrule
    \end{tabular}
  \end{minipage}
\end{table}

\section{Limitations}
\label{sec:limitations}
While ViDiT faithfully represents the semantic transformation encoded in the
input-edit pairs, the disentanglement of the learned direction is inherently bounded
by the disentanglement of the supervision source. When pairs exhibit residual
entanglement, the learned direction may reflect those characteristics. Additionally,
due to biases present in CLIP and Stable Diffusion, certain attributes such as age can
become entangled with correlated visual cues (e.g., white hair or eyeglasses). At
present, ViDiT optimizes a single global direction per concept, which may limit
expressiveness for highly localized or spatially varying edits where different regions
of the image require different degrees of modification. As the broader ecosystem of
editing methods continues to improve, ViDiT stands to directly benefit from
higher-quality supervision sources, and exploring this further remains an interesting
avenue for future work. Similarly, extending the framework to richer semantic concepts
beyond facial and artistic attributes such as object-level or scene-level edits
represents a natural and compelling direction for future investigation.

\section{Ethics Statement}
\label{sec:ethics}
ViDiT does not introduce any new bias of its own. It learns directions by transferring
semantics from pretrained models, including the supervision sources and the CLIP \cite{radford2021learning} and
Stable Diffusion \cite{rombach2022high} models it builds on, and therefore inherits whatever biases those models
already encode rather than creating additional ones. Any demographic correlations
observed in our edits, such as the residual cross-attribute shifts in our rescoring
analysis (Tab.~\ref{tab:rescoring}), originate in these pretrained components.

We anonymize racial attributes throughout: directions are referred to only by
non-semantic numeric labels (e.g., ``Race \#1'', ``Race \#2''), which identify
directions inherited from the source latent space and carry no ordering, ranking, or
preference for any group. We deliberately avoid explicitly naming any race, and our use
of these directions is intended purely to demonstrate transfer capability across diverse
semantics.

Finally, like all image synthesis and editing technologies, ViDiT carries a risk of
misuse for the generation of deceptive or manipulated media~\cite{korshunov2018deepfakes}.
We acknowledge this concern and encourage responsible deployment, including the
development of appropriate detection and attribution tools alongside such capabilities.

\section{Conclusion}
\label{sec:conclusion}
We introduced ViDiT, a framework that expands the editing vocabulary of pre-trained
diffusion models by transferring visual deltas directly from image-edit pairs into
continuous, transferable latent directions. Operating on a ``Learn Once, Edit Anywhere''
principle, ViDiT learns a single global direction from a small set of before-and-after
examples without fine-tuning the base model or requiring per-image optimization at
inference and applies it zero-shot to any image, including real photographs and
stylized artwork.

By framing the problem as visual-delta transfer rather than text-prompt engineering,
ViDiT overcomes the descriptive bottleneck of natural language and achieves
significantly stronger identity preservation than text-driven and instruction-following
baselines. The dual-objective training, combining semantic alignment via a frozen
vision-language encoder with latent alignment via the denoising network's native
prediction space, is key to achieving disentangled edits that preserve fine-grained
identity cues. Quantitative benchmarks and a user study with 40 participants confirm
that our zero-shot edits are perceived as more disentangled and semantically faithful
than existing alternatives.

While we leverage structured generative models as a high-quality supervision source,
the framework is entirely source-agnostic: directions learned from domain-adaptation
models, diffusion-based editors, or any method producing before-and-after image pairs
are equally well supported, pointing toward a general paradigm for incorporating
complex visual semantics into diffusion models from any generative source.

\section{Acknowledgements}
This research is supported by the National Science Foundation
under Grant No.\ 2543524.

\bibliographystyle{splncs04}
\bibliography{main}

\newpage

\appendix
\section{Qualitative Comparisons with Diffusion-based Editing Methods}
\label{sec:qualitative-supp}

To qualitatively compare the edits performed with diffusion-based editing methods and
the edits performed by ViDiT, we provide additional results here. In addition, we
further demonstrate the editing capabilities of our framework, such as edit strength
interpolation and generalization over out-of-domain images (e.g.\ images with style
components) to demonstrate how the edit strength can be controlled and applied to a
diverse set of imaging settings.

To compare the edits performed by ViDiT with diffusion-based editing methods, we
benchmark our approach against several recent approaches including Concept
Sliders~\cite{gandikota2023sliders}, SEGA~\cite{brack2023sega},
InstructPix2Pix~\cite{brooks2023instructpix2pix} and Prompt2Prompt
(P2P)~\cite{hertz2022prompt}. Here we use Null-text inversion (NTI) to adapt
Prompt2Prompt for real image editing (NTI + P2P). In particular, SEGA,
Prompt2Prompt, and InstructPix2Pix often result in substantial alterations to the
input image for edits like ``Race \#1''. In addition, Prompt2Prompt and SEGA lead to
entangled edits for edits such as ``Gender'', where alterations on attributes such as
age are also visible, whereas InstructPix2Pix produces results that are deteriorated
in terms of image quality. As illustrated in Fig. 5 of the main paper, ViDiT
surpasses these alternatives in maintaining semantic accuracy and in its ability to
execute disentangled edits. Concept Sliders face challenges in applying multiple edits
simultaneously, such as combining Race and Beard modifications, resulting in
significant deviations from the original input image, whereas ViDiT can successfully
combine multiple edits.

\begin{figure}[h]
    \centering
    \includegraphics[width=\linewidth]{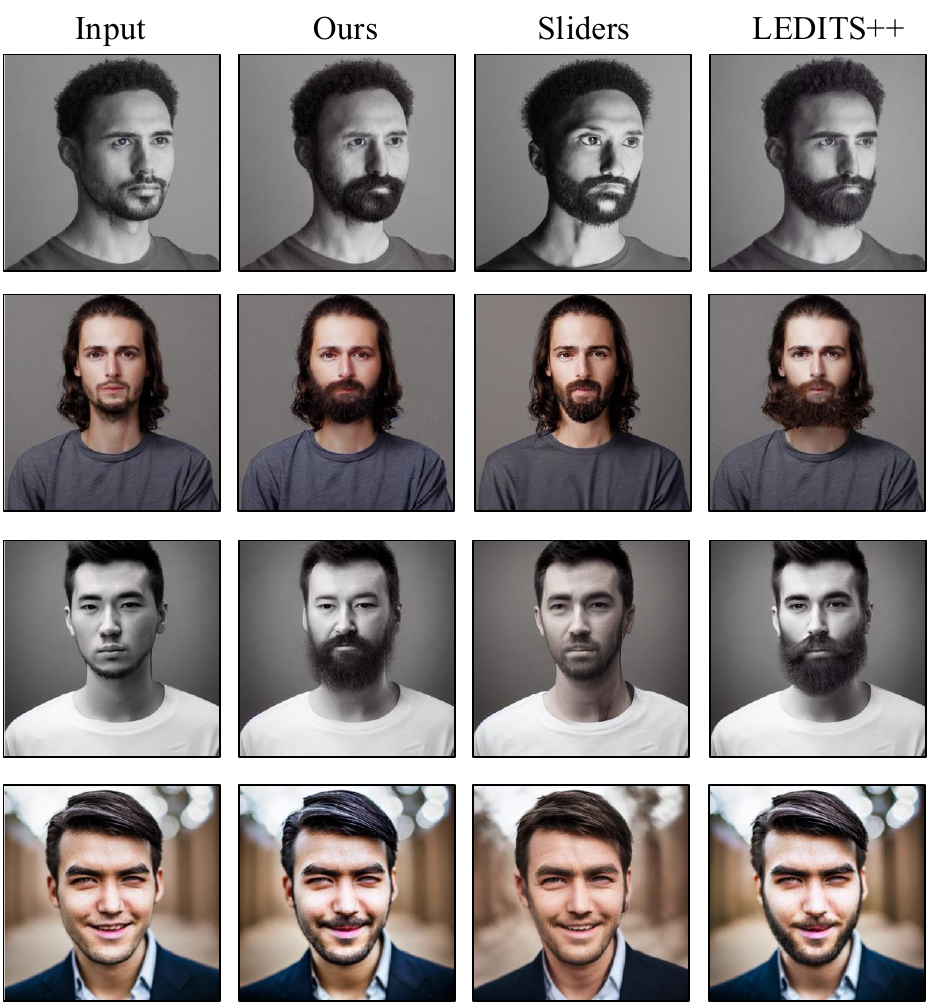}
    \caption{\textbf{Comparisons on Beard Edit.} We provide qualitative comparisons
    on the beard editing task with Concept Sliders~\cite{gandikota2023sliders} and
    LEDITS++~\cite{brack2023ledits++}, where we use image sliders
    in~\cite{gandikota2023sliders} for a fair comparison. ViDiT outperforms the
    competing approaches both in terms of editing quality and content preservation.
    Note that~\cite{gandikota2023sliders}, which learns the edit based on reference
    images, struggles when it attempts to add a beard to a sample without any traces
    of the attribute.}
    \label{fig:beard-comp}
\end{figure}

\begin{figure}[h]
    \centering
    \includegraphics[width=\linewidth]{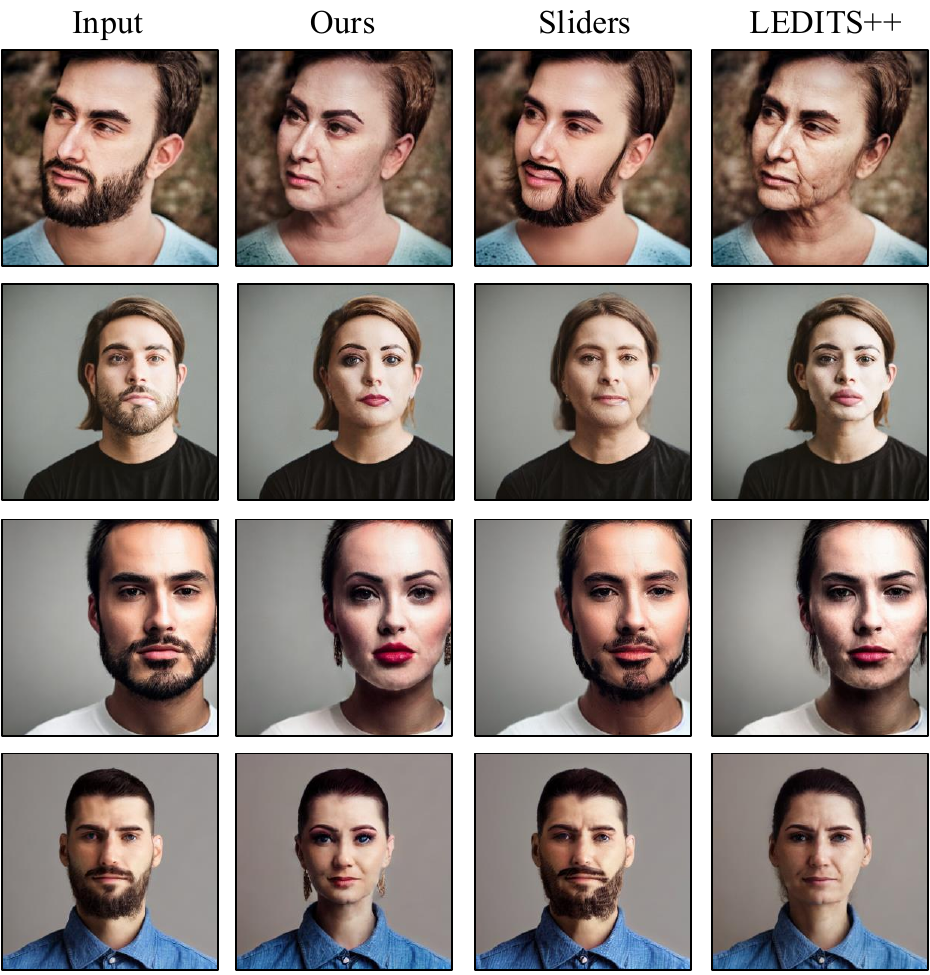}
    \caption{\textbf{Comparisons on Gender Edit.} To demonstrate the effectiveness
    of the gender editing direction learned by ViDiT, we provide qualitative
    comparisons with Concept Sliders~\cite{gandikota2023sliders} and
    LEDITS++~\cite{brack2023ledits++}. Notably, both~\cite{gandikota2023sliders}
    and~\cite{brack2023ledits++} struggle with artifacts while performing an edit
    that changes the overall appearance of the face, where~\cite{gandikota2023sliders}
    experiences it more severely. Directions learned by ViDiT perform such edits
    without sacrificing generation quality and in a disentangled manner.}
    \label{fig:gender-comp}
\end{figure}

\begin{figure}[h]
    \centering
    \includegraphics[width=\linewidth]{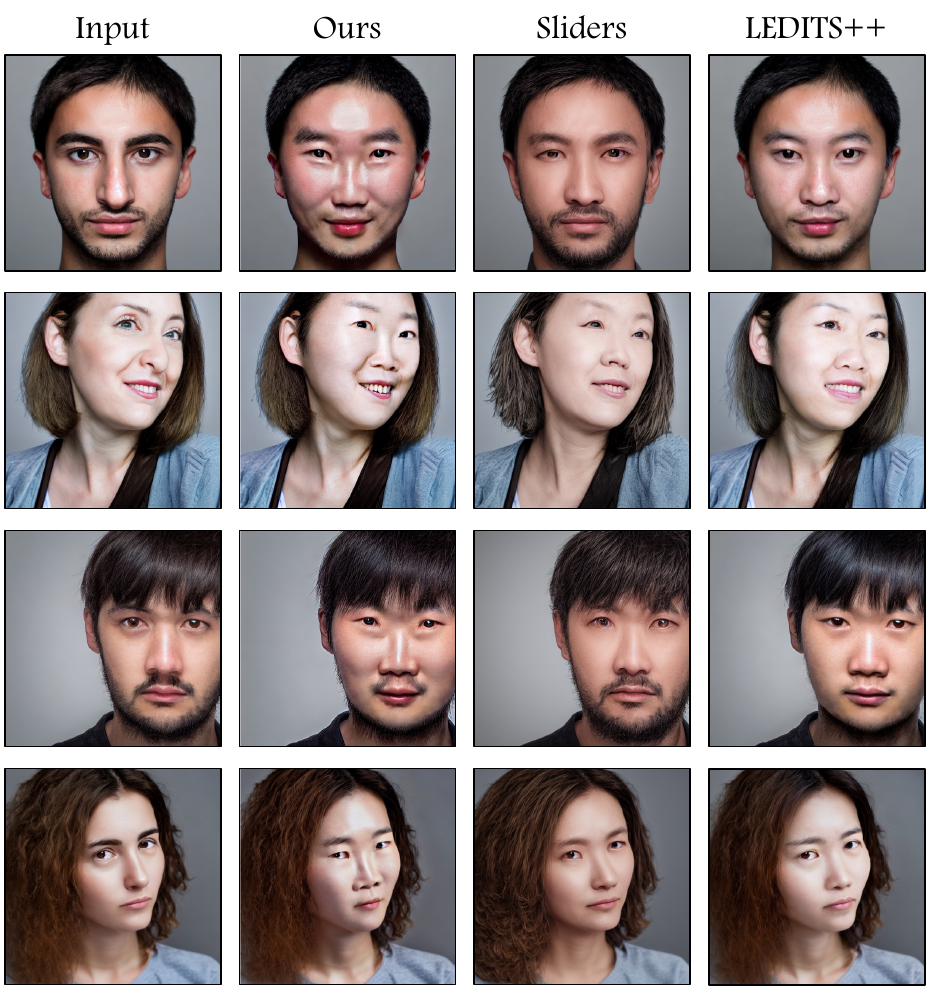}
    \caption{\textbf{Comparisons on Race \#1 Edit.} We provide qualitative comparisons
    on the attribute Race \#1 with Concept Sliders~\cite{gandikota2023sliders} and
    LEDITS++~\cite{brack2023ledits++}. Our method successfully reflects the edit while
    preserving the identity of the input image. Note that with LoRA-based approaches
    such as~\cite{gandikota2023sliders}, image quality is sacrificed in order to apply
    the edit, where significant changes to the input are present in the corresponding
    edits.}
    \label{fig:asian-comp}
\end{figure}

In addition, we provide comparisons with LEDITS++~\cite{brack2023ledits++}, which
requires text prompts to perform edits within the Stable Diffusion model. We perform
these comparisons on Beard, Gender, and Race semantics in
Figs.~\ref{fig:beard-comp}, \ref{fig:gender-comp} and~\ref{fig:asian-comp}. As can
be seen from the results, our method performs more disentangled edits compared to
LEDITS++. Furthermore, we provide comparisons with
Asyrp~\cite{kwon2022diffusion} and DiffAE~\cite{preechakul2022diffusion},
PnP-Diffusion~\cite{tumanyan2023plug} and MasaCtrl~\cite{cao2023masactrl} in
main paper Fig. 6.

\begin{figure}
    \centering
    \includegraphics[width=\linewidth]{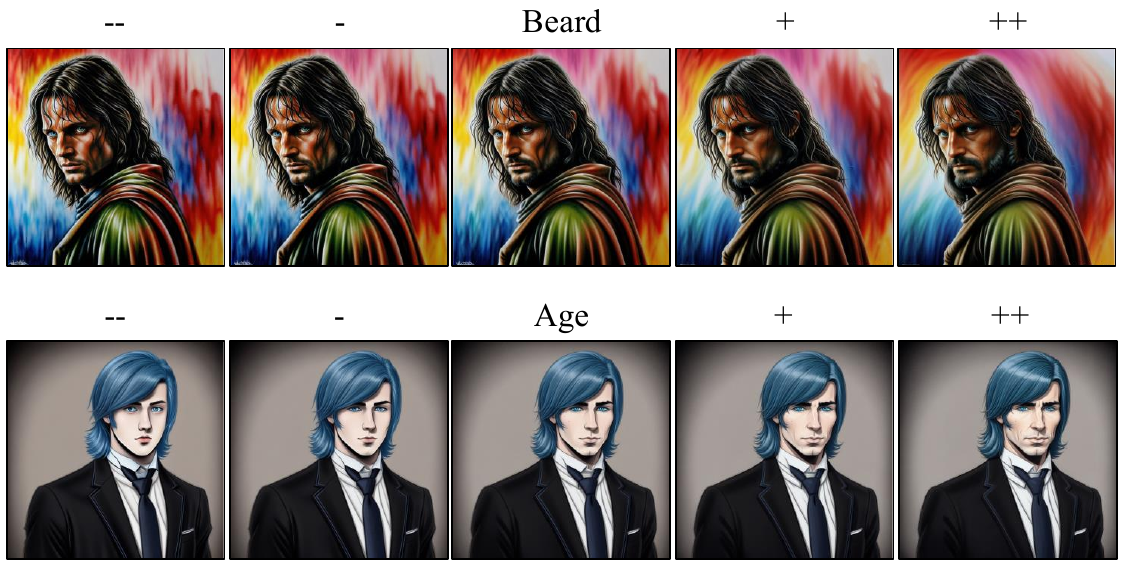}
    \caption{\textbf{Edit Interpolation Results on Generated Images.} We provide
    additional edit strength interpolation results over generated images for the
    attributes ``Age'' and ``Beard''. As demonstrated, the directions learned by
    ViDiT can be interpolated over images generated by Stable Diffusion, while
    preserving their stylization properties.}
    \label{fig:interpolate_gen}
\end{figure}

\section{Supplementary Editing Results and Details}

\noindent In addition to the results provided in the main paper, we provide additional
editing results in the supplementary material. Specifically, we provide edit
interpolation results on generated images in Fig.~\ref{fig:interpolate_gen}, editing
results on images generated by Stable Diffusion in Fig.~\ref{fig:supp_edit_gen},
editing results on artistic paintings in Fig.~\ref{fig:supp_edit_art}, and editing
results on real images in Fig.~\ref{fig:face_edit_supp}. Furthermore, we provide
additional details on the user study setup in Fig.~\ref{fig:user-study-setup}.

\section{Failure Cases}

\begin{figure}
    \centering
    \includegraphics[width=\linewidth]{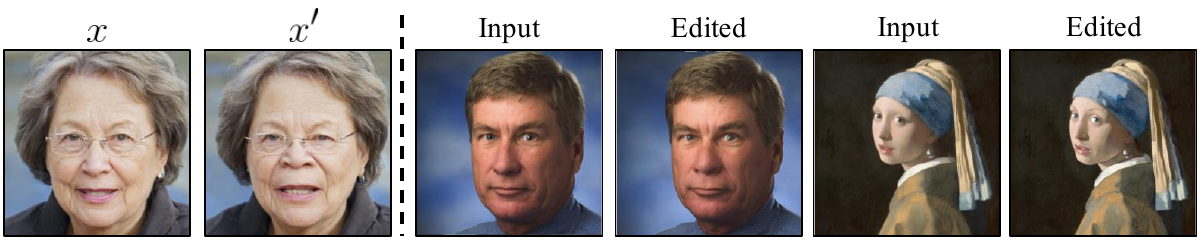}
    \caption{\textbf{Failure Cases of ViDiT.} When the supervision pairs encode
    subtle semantic changes (e.g.\ a mild smile), the learned direction $\mathbf{d}$
    may fail to capture sufficient signal to reproduce the edit faithfully on new
    images, as shown for a real photograph and an artistic painting. We attribute
    this to the limited semantic difference between $x$ and $x'$, which makes it
    difficult for the dual-objective training to isolate a meaningful direction.}
    \label{fig:limitations}
\end{figure}

\begin{figure}[h]
    \centering
    \includegraphics[width=\linewidth]{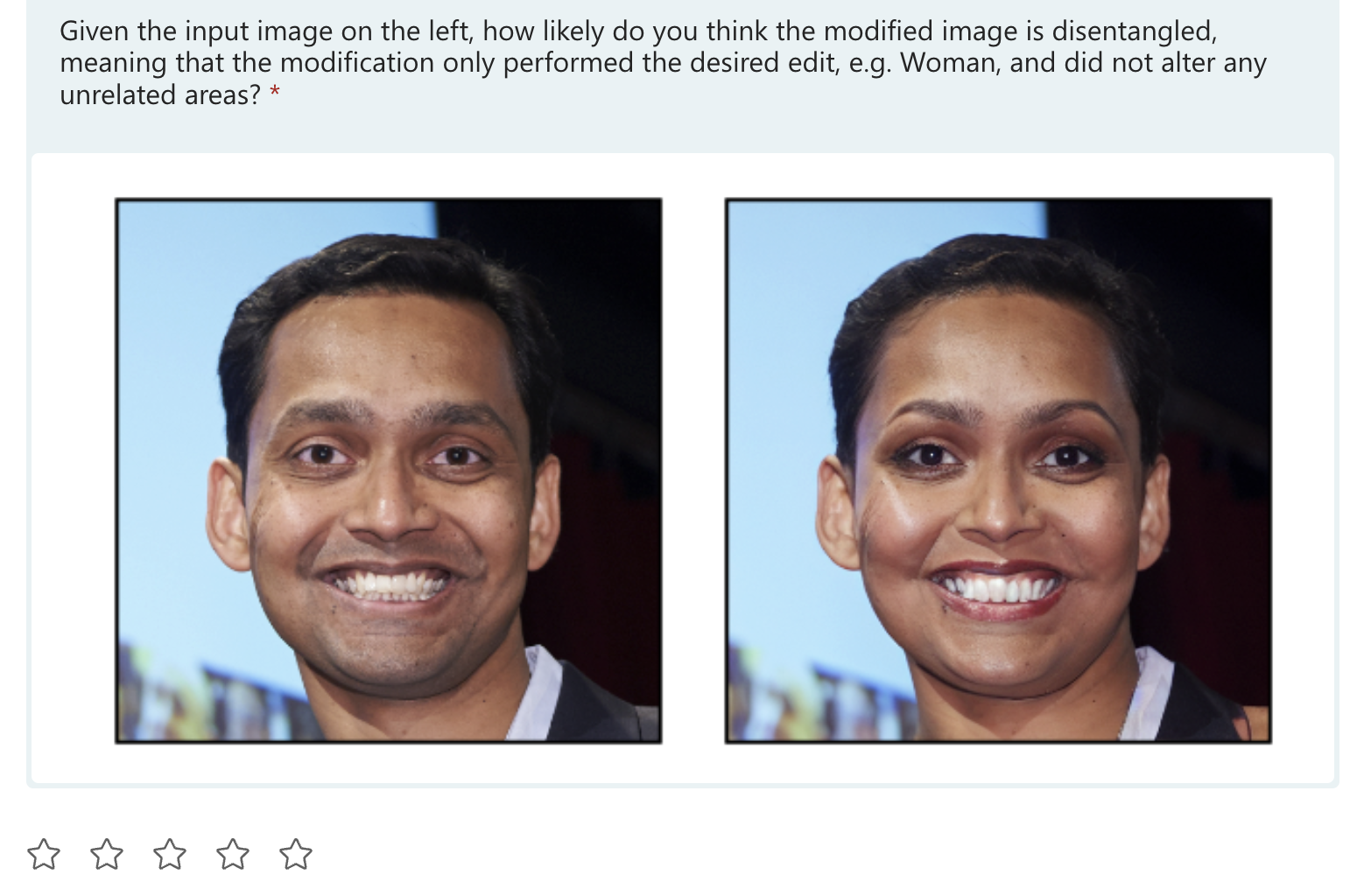}
    \caption{\textbf{User Study Setup.} To clarify the experimental setup used for
    the user study, we provide an example question. Users are shown an input-edit pair
    for a corresponding method and asked to assign a score from 1 to 5 based on their
    assessment of the edit.}
    \label{fig:user-study-setup}
\end{figure}

Extending our discussion on the limitations of ViDiT, we provide examples of failure
cases in Fig.~\ref{fig:limitations}. When the supervision
pairs encode very nuanced visual changes (e.g.\ minor modifications to nose shape),
the learned direction $\mathbf{d}$ may fail to capture sufficient semantic signal to
reproduce the edit faithfully. As demonstrated qualitatively, even in such cases ViDiT
can still localize the region to be edited; the limitation lies in the resolution of
detail that can be encoded within the diffusion model's conditioning space.

\section{Comparisons with \texorpdfstring{$\mathcal{W}+$}{W+} Adapter and Concept Sliders}
\label{sec:sup_discussion}

\begin{figure}[h]
    \centering
    \includegraphics[width=\linewidth]{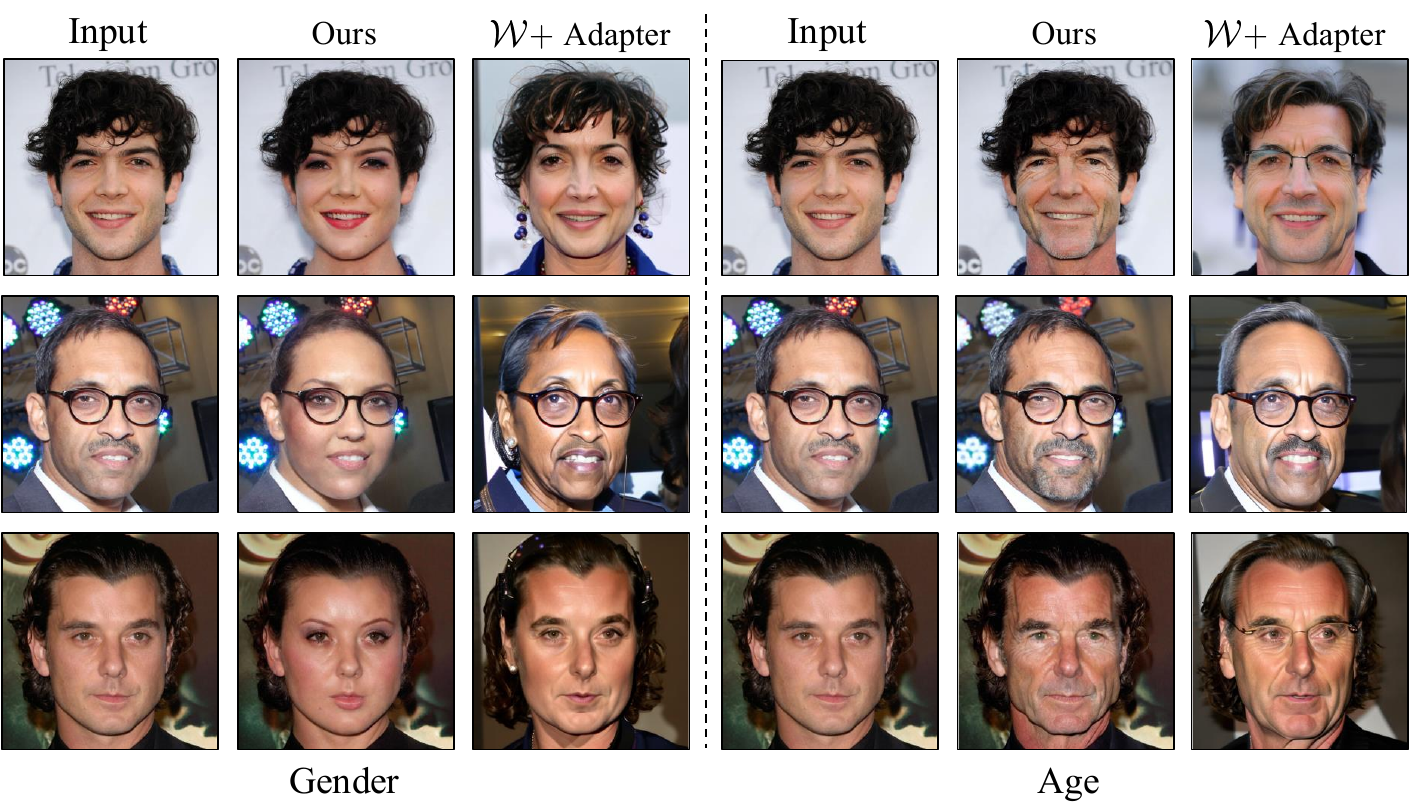}
    \caption{\textbf{Comparisons with $\mathcal{W}+$ Adapter~\cite{li2023w-plus-adapter}.}
    We compare ViDiT with~\cite{li2023w-plus-adapter} on the face editing task. Our
    method outperforms the adapter in terms of content preservation (preserving
    identity and details irrelevant to the edit) and disentangled editing (e.g.\
    disentangling attributes like eyeglasses and age).}
    \label{fig:w-plus-comparisons}
\end{figure}

We provide a qualitative comparison between the $\mathcal{W}+$
Adapter~\cite{li2023w-plus-adapter} and ViDiT in Fig.~\ref{fig:w-plus-comparisons}.
ViDiT learns continuous editing directions from before-and-after image pairs and
applies them to any given image without modifying the base model. In contrast, the
$\mathcal{W}+$ Adapter requires fine-tuning Stable Diffusion by training a separate
adapter module. Our results demonstrate that ViDiT more accurately preserves the
integrity of the input image while implementing the desired edits, such as changes in
Gender or Age.

Moreover, we provide qualitative comparisons with Concept
Sliders~\cite{gandikota2023sliders} in Figs.~\ref{fig:beard-comp},
\ref{fig:gender-comp} and~\ref{fig:asian-comp}, where ViDiT surpasses
\cite{gandikota2023sliders} both in disentanglement and representation quality without
the need to train separate LoRA models per direction.

\section{Supervision Pairs vs.\ Transferred Edits}
\label{sec:vs_stylegan}

\begin{figure*}[h]
    \centering
    \includegraphics[width=\linewidth]{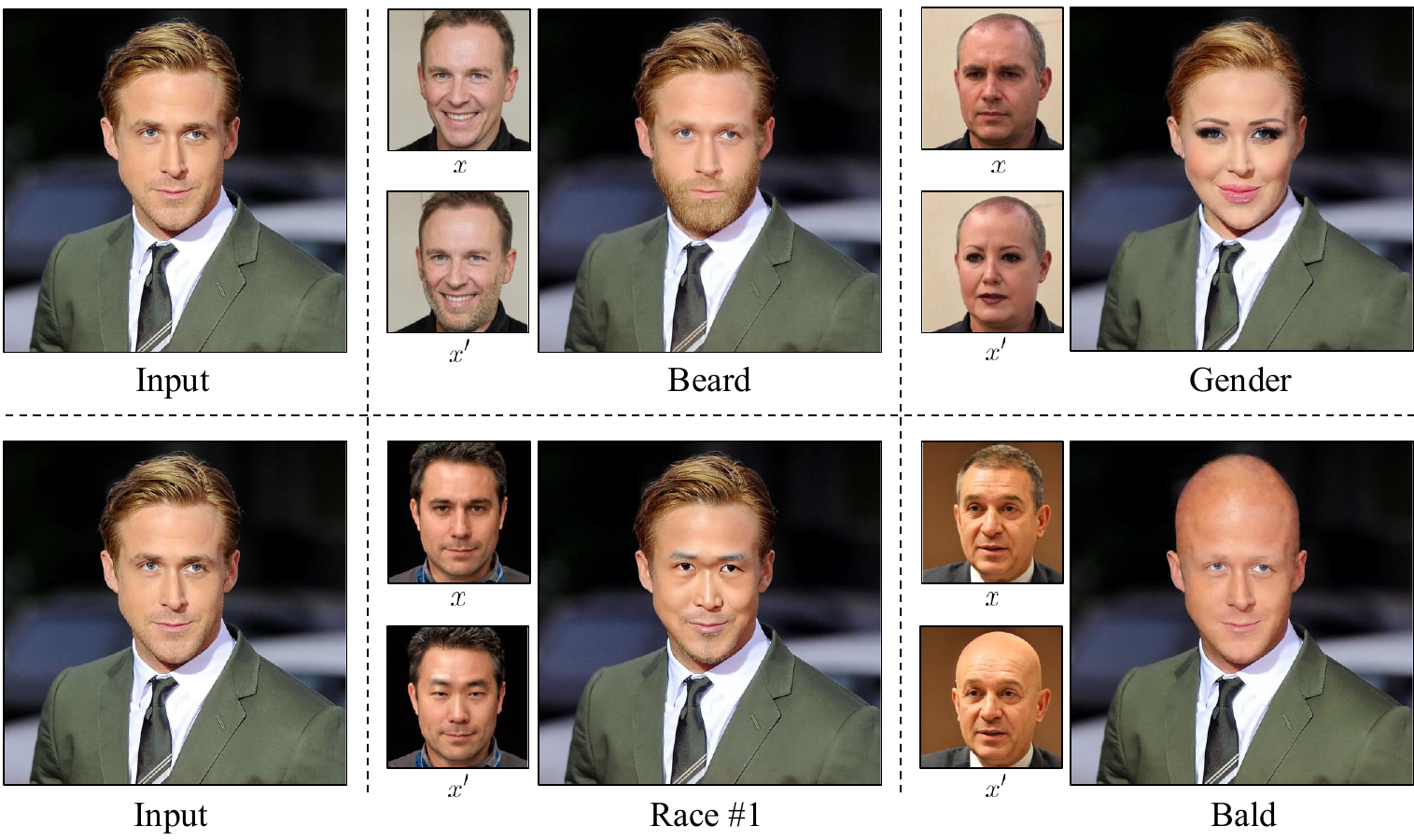}
    \caption{\textbf{Supervision Pairs vs.\ Directions Transferred by ViDiT.} We
    demonstrate Beard, Gender, Race \#1, and Baldness edits alongside the
    corresponding supervision pairs used during training ($x$ and $x'$,
    respectively). As the qualitative results show, ViDiT faithfully transfers the
    semantic transformations encoded in the pairs to new images via Stable Diffusion.}
    \label{fig:gan-to-diffusion}
\end{figure*}

\begin{figure*}
    \centering
    \includegraphics[width=\linewidth]{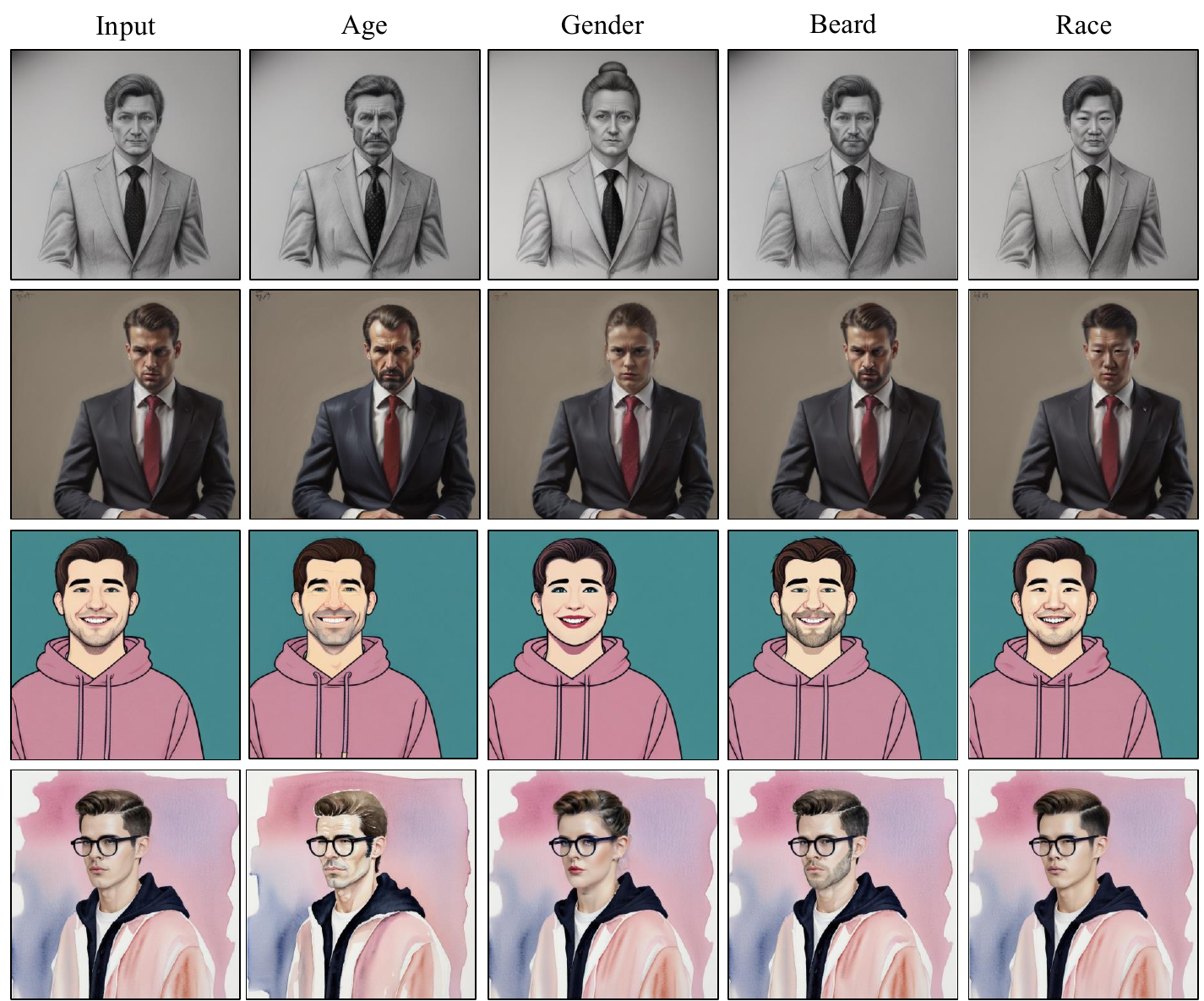}
    \caption{\textbf{Supplementary Editing Results on Images Generated by Stable
    Diffusion.} To show the generalization capabilities of ViDiT, we present
    qualitative results on images generated with Stable Diffusion, containing diverse
    style elements. Our directions apply the desired semantic without the need for
    inversion, while preserving the stylization characteristics of the edited images.}
    \label{fig:supp_edit_gen}
\end{figure*}

\begin{figure*}
    \centering
    \includegraphics[width=\linewidth]{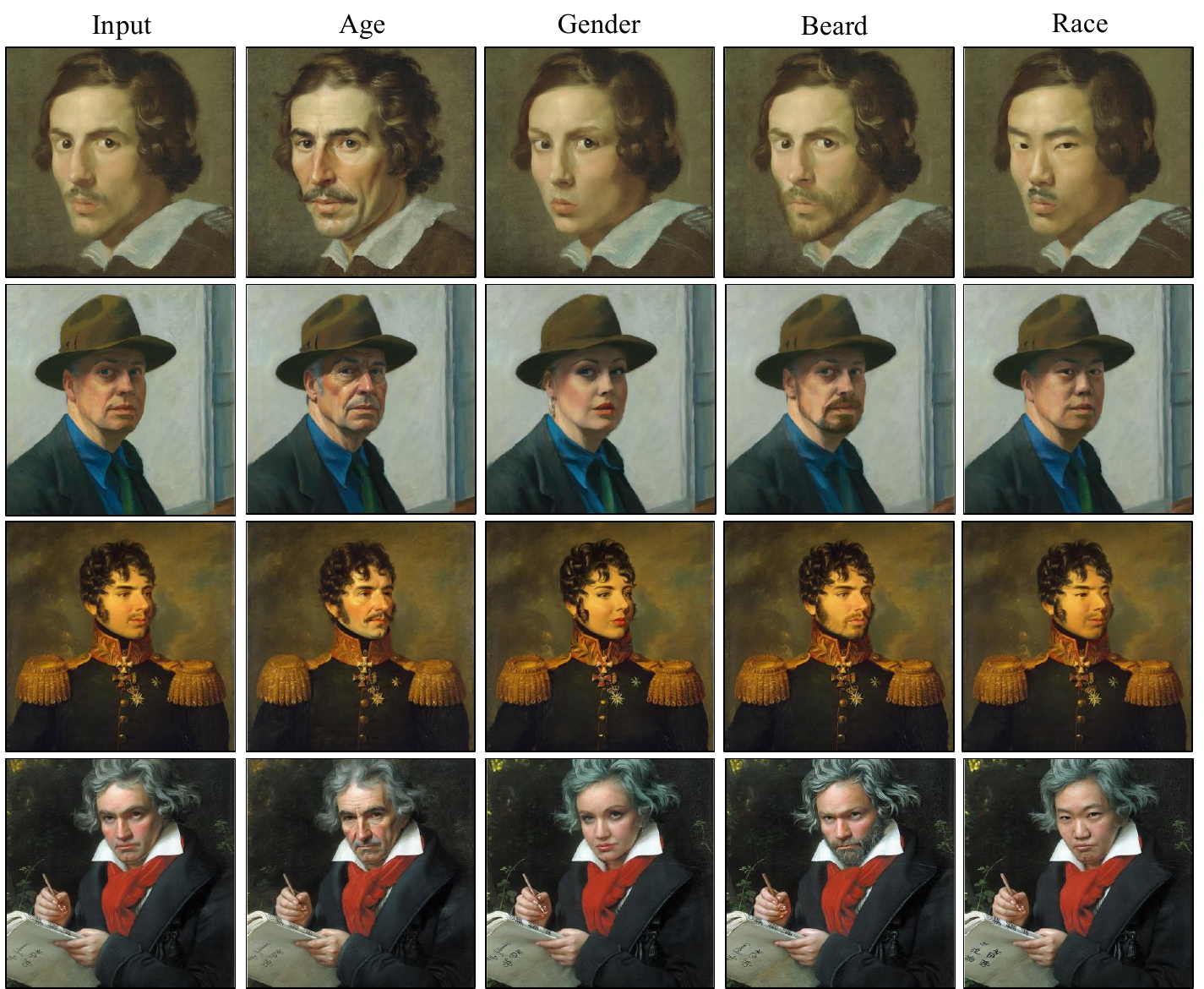}
    \caption{\textbf{Supplementary Editing Results on Artistic Paintings.} We provide
    additional editing results on painting images to further demonstrate compatibility
    with stylized content. As demonstrated qualitatively, ViDiT performs edits without
    altering the overall style of the image (e.g.\ oil painting).}
    \label{fig:supp_edit_art}
\end{figure*}

\begin{figure*}
    \centering
    \includegraphics[width=\linewidth]{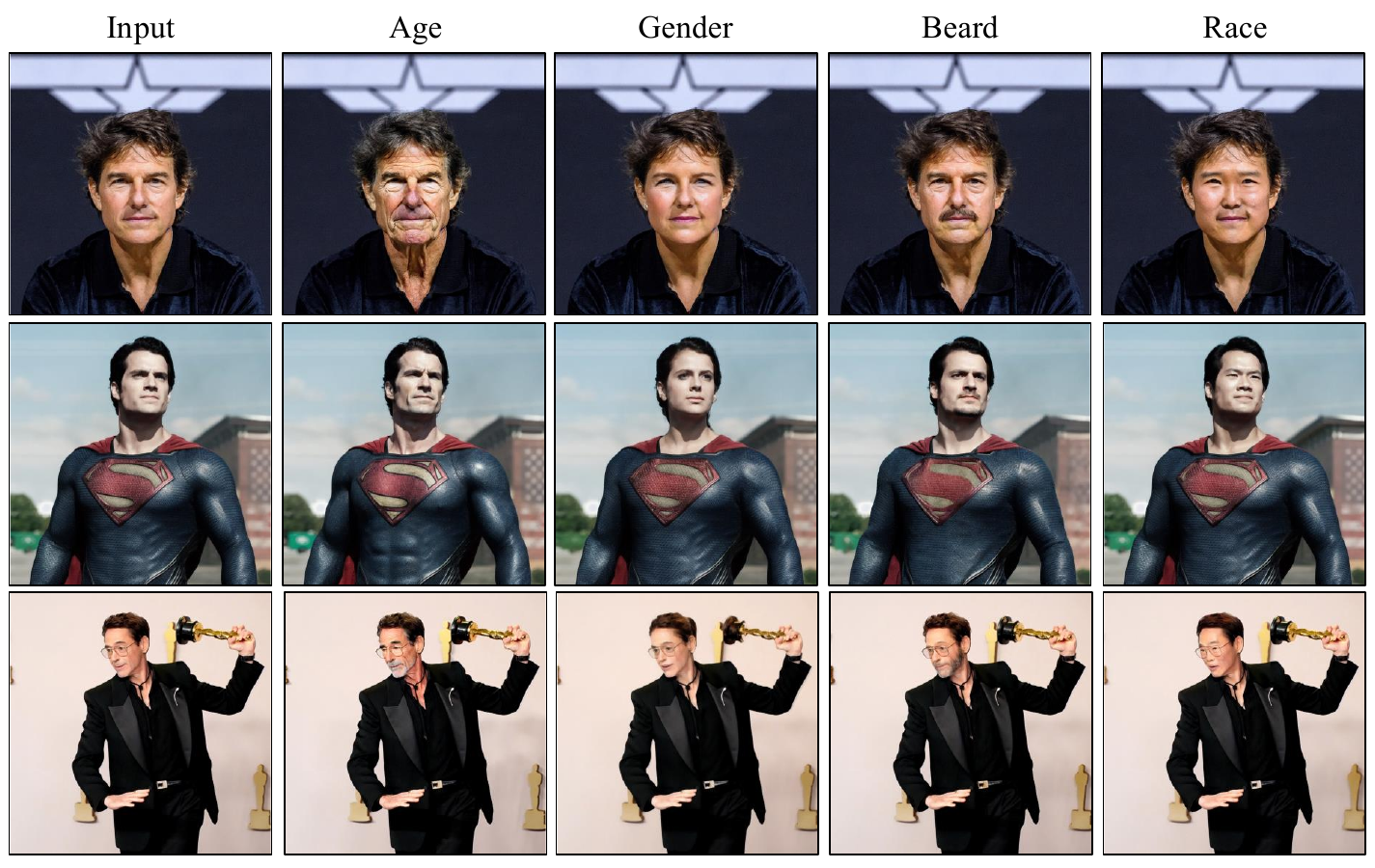}
    \caption{\textbf{Supplementary Editing Results on Real Images.} We provide
    additional editing results on real in-the-wild images, demonstrating that ViDiT
    generalizes beyond face-centered crops to diverse scenes and subjects while
    preserving identity and background.}
    \label{fig:face_edit_supp}
\end{figure*}

To illustrate how faithfully ViDiT transfers the semantic transformations encoded in
the supervision pairs, we show side-by-side comparisons between the input-edit pairs
used for training and the corresponding edits produced by ViDiT on new images in
Fig.~\ref{fig:gan-to-diffusion}.

\end{document}